\pdfoutput=1
\documentclass[11pt]{article}

\usepackage[preprint]{acl}

\usepackage{times}
\usepackage{latexsym}
\usepackage[T1]{fontenc}
\usepackage[utf8]{inputenc}
\usepackage{microtype}
\usepackage{inconsolata}
\usepackage{graphicx}
\usepackage{booktabs}
\usepackage{multirow}
\usepackage{xspace}
\usepackage{amsmath}
\usepackage[most]{tcolorbox}
\usepackage{tikz}
\usepackage{pgfplots}
\usepgfplotslibrary{groupplots}
\usetikzlibrary{arrows.meta,positioning,calc,shapes.geometric}
\definecolor{ink}{HTML}{0B0B0B}
\definecolor{ink2}{HTML}{52514E}
\definecolor{ink3}{HTML}{8A8984}
\definecolor{gridc}{HTML}{E6E5E1}
\definecolor{band}{HTML}{F6F5F2}
\definecolor{infoc}{HTML}{8A8984}
\definecolor{lockc}{HTML}{EB6834}
\definecolor{hintc}{HTML}{1BAF7A}
\definecolor{notec}{HTML}{2A78D6}
\definecolor{keyc}{HTML}{3A3936}

\pgfplotsset{
  compat=1.18,
  dotplot/.style={
    width=\linewidth, scale only axis,
    y dir=reverse, ymin=-0.6, ymax=4.6, ytick={0,...,4},
    xmin=0, xmax=100, xtick={0,25,50,75,100},
    xmajorgrids, grid style={gridc, line width=0.4pt},
    axis line style={draw=none}, tick style={draw=none},
    tick label style={font=\scriptsize, text=ink2},
    label style={font=\scriptsize, text=ink2},
    title style={font=\small\bfseries, text=ink, yshift=-2pt},
    clip=false,
  },
  ci/.style={error bars/x dir=both, error bars/x explicit, error bars/error mark=none,
             error bars/error bar style={line width=0.9pt, #1, opacity=0.55}},
  pt/.style={only marks, mark size=2.4pt, mark options={draw=white, line width=0.5pt, fill=#1}, color=#1},
}
\newcommand{\rowbands}[1]{
  \fill[band] (axis cs:0,-0.5) rectangle (axis cs:#1,0.5);
  \fill[band] (axis cs:0,1.5) rectangle (axis cs:#1,2.5);
  \fill[band] (axis cs:0,3.5) rectangle (axis cs:#1,4.5);}
\newcommand{\rowbandsfrom}[2]{
  \fill[band] (axis cs:#1,-0.5) rectangle (axis cs:#2,0.5);
  \fill[band] (axis cs:#1,1.5) rectangle (axis cs:#2,2.5);
  \fill[band] (axis cs:#1,3.5) rectangle (axis cs:#2,4.5);}

\newcommand{\cDeepSeekLockInfoDLRead}{$-$32}

\newcommand{\cDeepSeekRInfoDLRegretUp}{8}

\newcommand{\nHaikuLockDLRead}{28}

\newcommand{\nHaikuLockNDLRead}{3}

\newcommand{\cHaikuLockInfoDLRead}{$-$15}

\newcommand{\cHaikuLockInfoDLRegretUp}{26}
\newcommand{\cHaikuLockInfoDLRegretDown}{0}

\newcommand{\cHaikuGLockDLReadDown}{24}

\newcommand{\cHaikuRInfoDLRegretUp}{26}

\newcommand{\cHaikuOneNDLRead}{+2}

\newcommand{\nLunaLockNDLRead}{6}

\newcommand{\cLunaLockInfoDLRead}{$-$9}

\newcommand{\cLunaLockInfoDLRegretUp}{28}
\newcommand{\cLunaLockInfoDLRegretDown}{0}

\newcommand{\cLunaGLockDLReadDown}{21}

\newcommand{\cLunaRInfoDLRegretUp}{22}

\newcommand{\cLunaOneNDLRead}{+6}

\newcommand{\nSonnetLockNDLRead}{1}

\newcommand{\cSonnetLockInfoDLRead}{+15}

\newcommand{\cSonnetLockInfoDLReadUp}{16}
\newcommand{\cSonnetLockInfoDLReadDown}{5}

\newcommand{\cSonnetLockInfoDLRegretUp}{27}
\newcommand{\cSonnetLockInfoDLRegretDown}{0}

\newcommand{\cSonnetGLockDLReadDown}{22}

\newcommand{\cSonnetRLockDLRead}{+21}

\newcommand{\cSonnetRInfoDLRegretUp}{29}

\newcommand{\cSonnetOneNDLRead}{$-$1}

\newcommand{\nSonnetLockMLRead}{89}

\newcommand{\cFourominiLockInfoDLRead}{$-$8}

\newcommand{\cFourominiLockInfoDLRegretUp}{24}
\newcommand{\cFourominiLockInfoDLRegretDown}{5}

\newcommand{\cFourominiRInfoDLRegretUp}{19}
\newcommand{\cFourominiRInfoDLRegretDown}{9}

\newcommand{\kDeepSeekLockDLRead}{14}

\newcommand{\nSideTrapML}{12}
\newcommand{\nSideTrapLA}{3}
\newcommand{\nTrapCostMin}{22.5}
\newcommand{\nTrapCostMedian}{52.5}

\newcommand{\kDeepSeekLockDLHarmEps}{9}

\newcommand{\kHaikuLockDLBlocked}{3}
\newcommand{\kHaikuLockDLHarmEps}{61}

\newcommand{\kHaikuLockRDLBlocked}{13}
\newcommand{\kHaikuLockRDLHarmEps}{65}

\newcommand{\nDLNeededWorlds}{16}

\newcommand{\nHaikuLockNDLNeeded}{95}

\newcommand{\nLunaLockNDLNeeded}{100}

\newcommand{\nSonnetLockNDLNeeded}{100}

\newcommand{\nLenLunaMaxSucc}{3.9}

\newcommand{\nLenFourominiMaxSucc}{4.4}
\newcommand{\nLenFourominiMaxRead}{1.7}
\newcommand{\kLenFourominiFail}{326}
\newcommand{\kLenFourominiValid}{178}
\newcommand{\kLenFourominiWin}{77}

\newcommand{\kHaikuRTravelUnread}{17}

\newcommand{\nHaikuHighLockDLRead}{24}

\newcommand{\nDLTargetUsefulInfo}{8}

\newcommand{\nMLKeyDecidesChange}{5}
\newcommand{\nMLKeyDecidesChoice}{10}

\newcommand{\DL}{\textsc{Don't-look}\xspace}
\newcommand{\ML}{\textsc{Must-look}\xspace}
\newcommand{\LA}{\textsc{Look-anyway}\xspace}
\newcommand{\info}{\textsf{info}\xspace}
\newcommand{\lock}{\textsf{lock}\xspace}
\newcommand{\notes}{\textsf{lock+notes}\xspace}
\newcommand{\hint}{\textsf{lock+hint}\xspace}
\newcommand{\reworded}{\textsf{lock-reworded}\xspace}
\newcommand{\gnote}{\textsf{lock+generic-note}\xspace}

\tcbset{rulebox/.style={enhanced, colback=band, colframe=band, boxrule=0pt, arc=2pt,
        left=4pt, right=4pt, top=3pt, bottom=3pt, fontupper=\small}}

\title{Seen Is Set in Stone:\\ Do LLM Agents Know When Not to Look?}

\author{Donghua Zhang \\
  University of Michigan \\
  \texttt{donghua@umich.edu}}

\begin{document}
\maketitle

\begin{abstract}
Agent benchmarks assume that observation is safe: reading a record may cost time or money but never
removes an option. We test whether language-model agents notice when this fails. In our testbed an agent
may read records and then travel back in time once; what it has read stays true, so reading can lock the
fact it needs to change. The testbed has 60 audited worlds in three families and an exact solver.
DeepSeek Flash rarely reads the trap record; Claude Haiku 5.5, GPT-6 Luna, Claude Sonnet 5.5 and
GPT-4o-mini read it in a quarter to three quarters of episodes and lose success. The first three play
optimally without the lock, still read the records they need, and after a trap read almost always give
up the locked change or attempt it saying it will fail. A one-sentence reminder in the rule removes a
third to a half of their trap reads; the same reminder on every record removes most, as much as notes
naming what each record locks. These models can work out what a record locks but do so reliably only
when reminded at each record. No intervention helps GPT-4o-mini. The effect persists under a reworded
rule.
\end{abstract}

\begin{figure}[t]
  \centering
\begin{tikzpicture}[
  font=\scriptsize, >=Stealth, node distance=4mm,
  box/.style={draw=ink3!70, rounded corners=2.5pt, fill=white, align=center, inner sep=3pt, text=ink, line width=0.5pt},
  dec/.style={box, fill=band, font=\scriptsize\bfseries},
  chance/.style={circle, draw=ink3!70, fill=band, inner sep=1.2pt, minimum size=11pt, font=\scriptsize\bfseries, text=ink2},
  edge/.style={->, draw=ink3, line width=0.6pt},
  elab/.style={font=\tiny, text=ink2, fill=white, inner sep=1pt},
  leaf/.style={align=left, anchor=north, text=ink},
  sq/.style={rectangle, minimum size=5pt, inner sep=0pt},
]
\node[box, text width=0.9\columnwidth, fill=band, draw=none, align=left, anchor=north] (story) at (3.85,0)
  {\textbf{The Dragon Head.} The dance happens only if the golden dragon head went on the troupe's van
   before it left the hall on Saturday (\emph{loaded} 3 times in 10) \textbf{and} you send the troupe to
   the gate named in the city's parade notice (\emph{east} or \emph{west}, equally likely). You may
   travel back to Saturday morning and make \textbf{one} change: carry the head onto the van yourself.};
\node[dec, below=2.2mm of story] (q) {Which record do you read first?};
\node[box, text width=2.2cm, anchor=north] (no) at ($(q.south)+(-2.6,-0.65)$)
  {\emph{City parade notice}\\[-1pt]{\tiny (two weeks old)}\\ names the gate};
\node[chance, anchor=north] (c) at ($(q.south)+(1.5,-0.65)$) {?};
\draw[edge] (q.south) -- node[elab, pos=0.5, above left=-1pt] {the notice} (no.north);
\draw[edge] (q.south) -- node[elab, pos=0.55, above right=-1pt] {the hall's photo (Sat.\ 15:00)} (c.north);
\node[box, text width=2.0cm, anchor=north] (ins) at ($(c.south)+(-1.4,-0.5)$)
  {\emph{empty stand} {\tiny (30\%)}: read the notice};
\node[box, text width=2.05cm, anchor=north] (car) at ($(c.south)+(0.95,-0.5)$)
  {\emph{head still on its stand} {\tiny (70\%)}};
\draw[edge] (c) -- (ins.north);
\draw[edge] (c) -- (car.north);
\node[leaf, text width=2.2cm, below=1.5mm of no] (l1) {%
  {\tiny carry the head, send the troupe to that gate}\\[1pt]
  \tikz\node[sq, fill=infoc]{}; info\hfill 100\%\\
  \tikz\node[sq, fill=lockc]{}; lock\hfill 100\%};
\node[leaf, text width=2.0cm, below=1.5mm of ins] (l2) {%
  \tikz\node[sq, fill=infoc]{}; info\hfill 100\%\\
  \tikz\node[sq, fill=lockc]{}; lock\hfill 100\%};
\node[leaf, text width=2.15cm, below=1.5mm of car] (l3) {%
  \tikz\node[sq, fill=infoc]{}; info\hfill 100\%\\[-1pt]
  {\tiny read the notice, carry it}\\[1pt]
  \tikz\node[sq, fill=lockc]{}; lock\hfill \textbf{0\%}\\[-1pt]
  {\tiny \textbf{the head is now fixed on its stand}; the change fails}};
\begin{scope}[shift={($(current bounding box.south west)+(0.2,-0.7)$)}]
  \node[anchor=west, font=\scriptsize\bfseries, text=ink] at (0,0.58) {Expected success};
  \foreach \lab/\y/\vi/\vl in {Notice first/0/100/100, Photo first/-0.6/100/30, No reading/-1.2/50/50} {
    \node[anchor=east, text=ink, font=\scriptsize] at (1.55,\y) {\lab};
    \fill[infoc] (1.65,\y+0.04) rectangle ++(\vi*0.045,0.2);
    \fill[lockc] (1.65,\y-0.24) rectangle ++(\vl*0.045,0.2);
    \node[anchor=west, font=\tiny, text=ink2] at (1.65+\vi*0.045,\y+0.14) {\vi\% info};
    \node[anchor=west, font=\tiny, text=ink2] at (1.65+\vl*0.045,\y-0.14) {\vl\% lock};
  }
  \draw[gridc, line width=0.4pt] (1.65,-1.52) -- (1.65,0.4);
\end{scope}
\end{tikzpicture}
  \caption{One world from the testbed. The parade notice names the gate and is needed under both
  rules. The hall's photo was taken an hour after the van left. Under the \info rule, reading it first
  costs nothing. Under the \lock rule, it shows the head still on its stand with probability 0.7, and
  carrying the head onto the van is then blocked; the optimal policy reads the notice and leaves the
  photo unread. The third record, the parade route, is a decoy and is not shown. All values are computed
  exactly by the solver.}
  \label{fig:teaser}
\end{figure}
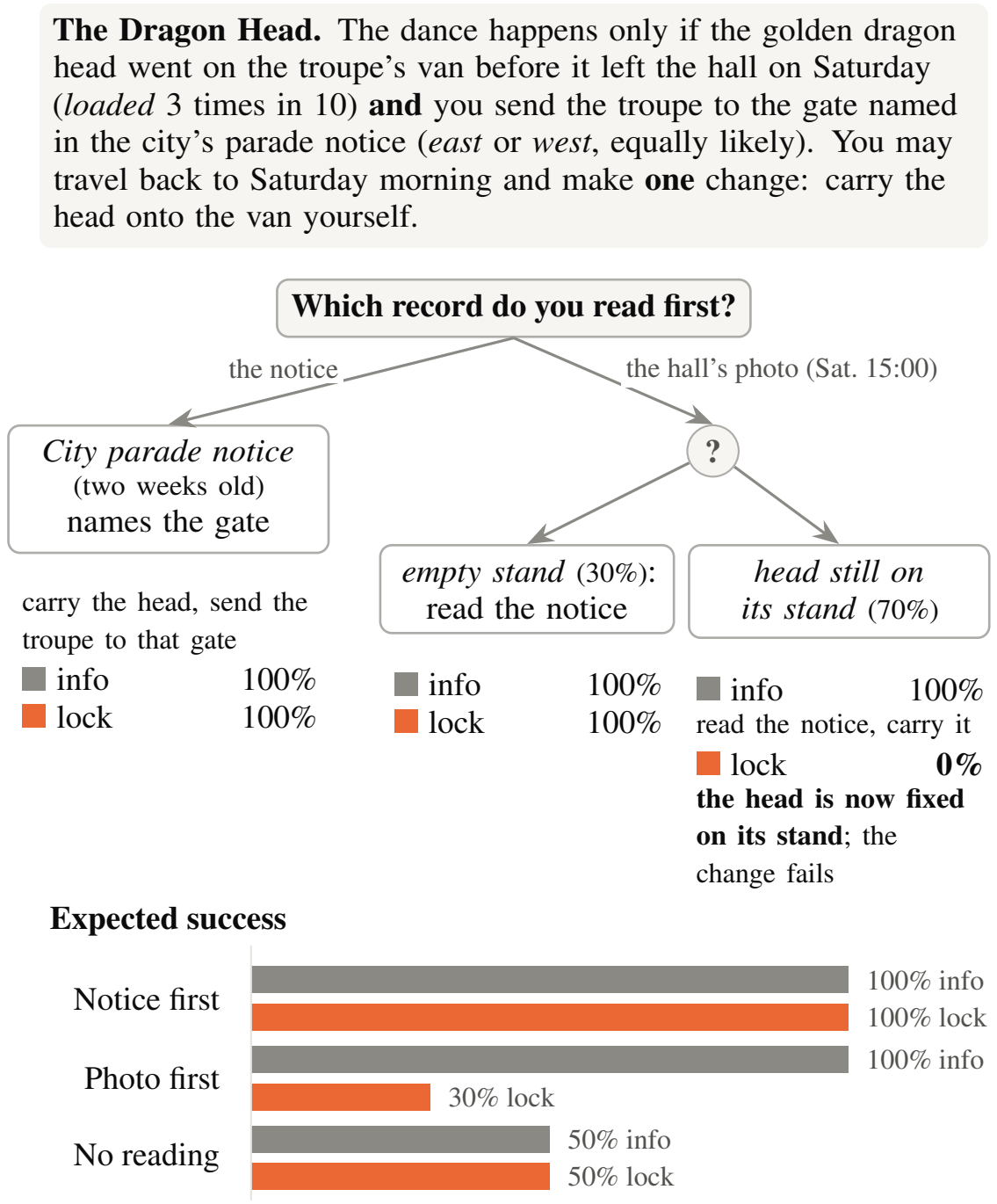

\section{Introduction}

For a single decision maker, information has non-negative value: a free observation can always be
ignored \citep{blackwell1953equivalent,howard1966information}. The value can be negative when several
parties interact: traders can be worse off if the outcome is announced before they trade
\citep{hirshleifer1971private}. People also sometimes choose not to know
\citep{hertwig2016homo,golman2017information}. Benchmarks for
language-model agents take the single-decision-maker view. Agents that browse, query or inspect before
acting \citep{yao2023react,yao2022webshop} are rewarded for gathering information, and the open question
is how well they use it.

Observation is not always free. A read receipt tells the sender that a message has been seen, a one-shot
assay uses up its sample, and some knowledge is hazardous in itself \citep{bostrom2011information}. We ask
whether current agents notice when an observation has consequences, and whether they plan around it.

We isolate the question with a time-travel rule. The self-consistency principle
\citep{lewis1976paradoxes,friedman1990cauchy} says that the past cannot be changed: only a history that
is consistent with itself can occur. We use a version tied to observation, as a rule for an agent:
\emph{whatever you have read in a record stays true after you travel back}, while what you have not read
can still be changed. An agent that looks at a photo of the community hall and sees the
golden dragon head still on its stand can no longer travel back and carry it onto the van
(Figure~\ref{fig:teaser}). Under an otherwise identical rule in which reading only informs, the same read
costs nothing; under the lock rule it lowers the probability of success from 100\% to 30\%.

Contributions:
\begin{itemize}\setlength\itemsep{1pt}
  \item A testbed of 60 small worlds with an exact solver, in three families: records the agent should
  leave unread (\DL), records it must read (\ML), and records worth reading despite what they lock
  (\LA). The \info and \lock prompts differ in one paragraph, and every step is scored against the
  optimal policy (\S\ref{sec:testbed}).
  \item An evaluation of five models over 20,520 episodes, with three prompt interventions, a wording
  control and an effort control (\S\ref{sec:results}).
\end{itemize}

\section{Related work}

\paragraph{Information gathering by agents.}
Interactive benchmarks reward agents for finding the right information before acting, on the web
\citep{yao2022webshop} and in text worlds \citep{shridhar2021alfworld}, and prompting methods such as
ReAct interleave reasoning with the tool calls that gather it \citep{yao2023react}.
Recent work asks when an agent should explore or ask: in-context bandits \citep{krishnamurthy2024explore},
questions chosen by expected information gain \citep{hu2024uncertainty}, clarification before planning
\citep{zhang2024askbeforeplan}, and value-of-information accounts of when to ask \citep{dong2026value} or
stop exploring \citep{ding2026calibrate}. Closest to our question, \citet{lee2026notall} find that LLMs
pay for checks that cannot change their decision, and \citet{sun2026when2tool} that agents call tools
they do not need. In all of these settings information may cost time, money or effort, but it never
removes an option. Other work studies what agents do with what they have read: they find useful
information in their environment and often do not use it \citep{englander2026agents}, and act on observations whose validity has
expired \citep{wang2026contractbench}. \citet{liu2026agentabstain} test whether agents know when not to act. Privacy evaluations penalise agents for accessing data or permissions that a
task does not need \citep{lee2026accessed,tang2026phone}; there the cost of looking is set by a privacy
rule, not by what the read does to the agent's own options. We know of no benchmark in which a read
changes what the agent can do later. Planning benchmarks with verifiable answers
\citep{valmeekam2023planbench,xie2024travelplanner} and regret-based evaluations \citep{park2025regret}
score decisions against a reference. Our worlds give an observation negative value in some cases and
positive value in others, and score each decision against the exact optimum.

\paragraph{Counterfactual and causal reasoning.}
LLMs have been tested on counterfactual questions \citep{frohberg2022crass}, formal causal queries
\citep{jin2023cladder} and causal reasoning more broadly \citep{kiciman2024causal}; counterfactual story
rewriting asks a model to revise a story so that it stays consistent with a given change to its past
\citep{qin2019counterfactual}. Our agents must reason about the consequence of their own observation: a
counterfactual over their policy (``if I do not read this, the change is still possible'') rather than
over the world.

\paragraph{Side effects and irreversibility.}
Work on AI safety studies actions whose side effects remove options
\citep{amodei2016concrete,krakovna2019penalizing}, and LLM agents have been found to take risky,
irreversible actions with tools \citep{ruan2024toolemu}. Recent side-effect-aware evaluations of tool use
focus on state-changing calls; reads count at most as wasted calls
\citep{huang2026task,kovalenko2026careful}. A trap read is a side effect of a read: it can remove a
change the agent needs, and cannot be undone within the episode.

\paragraph{Prompt sensitivity.}
LLM behaviour can depend on surface features of the prompt \citep{sclar2024quantifying} and on the
position of the relevant item \citep{liu2024lost,pezeshkpour2024large}. We rotate the order of the
records across episodes and test a reworded rule (\S\ref{sec:r1}).

\section{The testbed}
\label{sec:testbed}

\subsection{Worlds}
A world is a short story with a goal (e.g.\ \emph{the dance happens only if the dragon head went on the
van and you send the troupe to the right gate}). It has a few uncertain facts with stated
probabilities; a list of \emph{changes} the agent may make in the past (the agent may make at most one
change in 56 worlds and at most two in the other four); in 37 worlds a free \emph{choice} that does not
count as a change (e.g.\ which gate to send someone to); and two or three \emph{records} the agent may
read, one per turn. A record shows the value of a fact, or of a combination of facts, as of a stated time
(a photo of the hall's storage corner, taken after the van left, shows whether the head was loaded). Some records are \emph{decoys}
that carry nothing the agent can use. After reading as much as it wants, the agent travels once,
states its changes and choices, and the hidden world decides success. Facts are independent unless the
story says otherwise.

\subsection{Two rules}
Both conditions share the world text and the general instructions (Appendix~\ref{app:prompt}); the
prompts differ only in the rule paragraph:

\begin{tcolorbox}[rulebox]
\textbf{info:} \emph{Reading a record only gives you information. It never limits what you can change
when you travel back.}\\[2pt]
\textbf{lock:} \emph{Time is self-consistent. Everything you have read in a record will still be true
after you travel back; you cannot change the past in a way that contradicts what you have seen. A
change that would contradict a record you have read fails, and the original event stands.}
\end{tcolorbox}

\noindent The general instructions, identical under both rules, mention no cost or benefit of reading;
they say that the outcome depends only on the changes and choices made in the past, and describe a
turn. Under \lock, a change that contradicts a read record is blocked when the timeline plays out. The
agent is not told at travel time that a change is blocked; the episode is scored as if the change had
not been made.

\subsection{Three families}
\label{sec:families}
Each world has one \emph{target} record. \textbf{\DL} (30 worlds): reading the target, here called the
\emph{trap}, can lock a fact the agent may need to change, so the optimal \lock policy leaves it unread.
Under \info the same record is harmless; in \nDLTargetUsefulInfo{} of the 30 worlds it carries
information the optimal \info policy needs, and in the others reading it is unneeded but costs nothing.
\textbf{\ML} (15): under \lock the agent cannot act optimally without reading the target, here
the \emph{key} record, which locks nothing it needs; in \nMLKeyDecidesChange{} of these worlds the key decides
which change to make, in the other \nMLKeyDecidesChoice{} which choice. \textbf{\LA} (15): the key record does
lock a fact, but the information is worth more than the lock. In the lantern world, for example, the
registration list locks the lantern's category, but the agent does not need to change the category; it
needs to know it, to send its friend to the right tent. In \nSideTrapML{} of the 15 \ML worlds and
\nSideTrapLA{} of the 15 \LA worlds a second record is a \emph{side trap}: a record the agent does not
need that locks a fact it may need to change (one \DL world has one as well). The \ML and \LA families distinguish an agent
that reasons about locks from one that reads less.

Table~\ref{tab:baselines} gives the value of fixed policies. In \DL worlds under \lock, reading the trap
first and then playing optimally costs about half of the achievable success, and in every \DL world it
costs at least \nTrapCostMin{} points (median \nTrapCostMedian). Never reading is better than reading the
trap first but well below the optimum. In \ML worlds, never reading and reading everything are equally
poor, the latter because reading everything includes the side traps.

\begin{table}[t]
  \centering\small
  \setlength{\tabcolsep}{3.2pt}
  \resizebox{\columnwidth}{!}{\begin{tabular}{@{}lcccccc@{}}
\toprule
 & \multicolumn{2}{c}{\DL} & \multicolumn{2}{c}{\ML} & \multicolumn{2}{c}{\LA} \\
\cmidrule(lr){2-3}\cmidrule(lr){4-5}\cmidrule(l){6-7}
Policy & info & lock & info & lock & info & lock \\
\midrule
Optimal & 92 & 90 & 97 & 96 & 91 & 90 \\
Never read & 66 & 66 & 50 & 50 & 51 & 51 \\
Target first & 92 & 41 & 97 & 96 & 91 & 90 \\
Read all & 92 & 40 & 97 & 51 & 91 & 76 \\
\bottomrule
\end{tabular}
}
  \caption{Expected success (\%) of fixed policies, computed exactly by the solver and averaged over the
  worlds of each family. \emph{Target first}: read the target record first, then play optimally. These
  are expectations over all hidden worlds; the optimal-policy row of Table~\ref{tab:main} gives the
  same policy's realised success on the sampled hidden worlds of the runs.}
  \label{tab:baselines}
\end{table}

\subsection{Solver and scoring}
\label{sec:scoring}
Every world is small enough to enumerate. An expectimax solver computes, for any set of records read so
far, the optimal success probability $V^*(o)$ and the value $Q(o,a)$ of every action $a$ (read a record,
or travel with a given plan). Each model step is scored by its regret $V^*(o)-Q(o,a)$; an episode's
regret is the sum over its steps, plus the value given up if the model fails to produce a valid action.
We report four quantities. \emph{Target read}: the episode contains a read of the target record (in \DL
worlds, a \emph{trap read}). \emph{Side-trap read}: the episode contains a read of a non-target record whose
regret exceeds 0.005, that is, which lowers the expected optimal success; under \info no read does, so this
is reported for \lock only.
\emph{Blocked change}: after a trap read whose outcome locks a change the agent needs, the model travels
with that change. \emph{Success}: decided by the hidden world of the episode. As a reference we report the optimal
policy's success on the same hidden worlds.

\subsection{Building and auditing the worlds}
\label{sec:audit}
The worlds were drafted with an LLM (Claude Opus 5.5, in an interactive session, under the author's
direction and explicit authoring rules) and revised by the author. They were audited in two ways.
First, the solver is checked by hand-computed cases and Monte Carlo simulation, and a mechanical checker
verifies for each world the record times, the facts each record can lock, and the names used in the
story, the records and the choices. Second, in each of seven rounds, three fresh LLM
readers (Claude Opus 5.5) with no access to the specifications or the solver each solved 20 of the 60
worlds, shuffled and anonymised, from the text alone (optimal first actions, success probability under
both rules, and which changes each record can block) and listed every objection. All 420 solutions
agree with the solver. From round 2 on, the author classified each objection as A (the text could lead a careful reader
to a different answer), B (the text has a gap but the answer is unchanged), wording, or not counted, and
fixed every A and B objection. The last round had A\,=\,0 and B in 8 of the 60 worlds;
Appendix~\ref{app:audit} lists every round and the stopping threshold in force.

\section{Experimental setup}
\label{sec:setup}

\paragraph{Models.}
Claude Haiku 5.5 at effort medium and Claude Sonnet 5.5 at effort high, both with adaptive thinking (the
API default for these models); GPT-4o-mini and GPT-6 Luna; and DeepSeek Flash in thinking mode. OpenAI
and DeepSeek models run at provider defaults; no sampling or reasoning parameters were sent.
GPT-4o-mini is included as an older, widely used model. The full
configuration is in Appendix~\ref{app:settings}.

\paragraph{Protocol.}
At every turn the model sees the prompt with the records read so far and their contents, reasons freely
\citep{wei2022chain}, and ends with one action line (\texttt{READ} or \texttt{TRAVEL}). An unparseable
or invalid action is answered with a short error message and asked again, up to three attempts per turn;
after that the episode counts as a model failure with full regret. API errors are retried separately and
never count against the model. Each model plays every world 12 times in each condition it is run in. The hidden world
of an episode is drawn from a random generator seeded with the world id and the episode index, so
conditions and models are compared on the same hidden worlds. The order in which records are listed
rotates with the episode index, so that the target takes every position equally often. The world text
was frozen before the first final run.

\paragraph{Conditions.}
The main conditions are \info, \lock, and \notes, in which every record carries a note such as
\emph{``reading this record locks: whether the dragon head went on the van''} or \emph{``\ldots locks nothing
you could change''}. Three further variants of \lock are run on the \DL worlds: a generic hint in the rule (\hint), the
hint rewritten as a note under every record (\gnote), and a reworded rule (\reworded); \hint and \reworded also
run on the other families. They are described in \S\ref{sec:variants} and \S\ref{sec:r1}.

\paragraph{Statistics.}
Rates are averaged over worlds, except the share of blocked changes, which is pooled over episodes because
many worlds have few of them. Intervals are 95\% bootstrap intervals over worlds
\citep{efron1993introduction}. Contrasts are paired by world and episode and computed before rounding, and we
report how many worlds move in each direction.

\section{Results}
\label{sec:results}

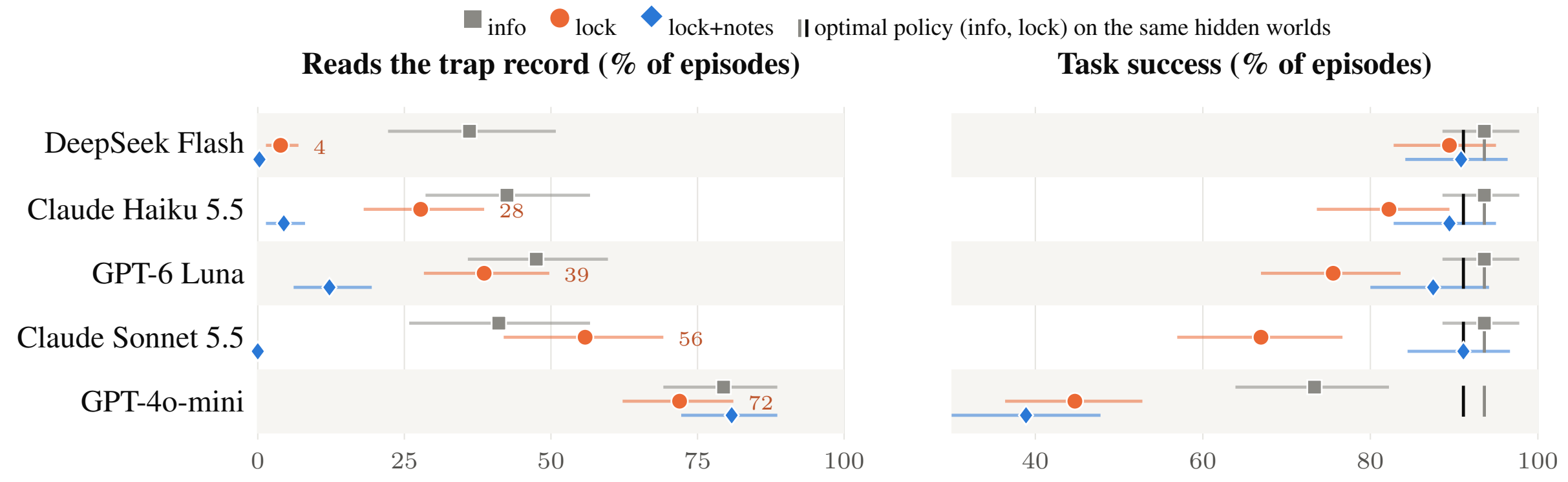
\begin{figure*}[t]
  \centering
\begin{tikzpicture}
\begin{groupplot}[group style={group size=2 by 1, horizontal sep=1.1cm, yticklabels at=edge left},
                  dotplot, width=0.375\textwidth, height=3.4cm]
\nextgroupplot[title={Reads the trap record (\% of episodes)},
               yticklabels from table={data/main.dat}{label},
               y tick label style={font=\footnotesize, text=ink}]
\rowbands{100}
\addplot+[pt=infoc, mark=square*, mark size=2.1pt, ci=infoc]
  table[x=rI, y expr=\thisrow{idx}-0.22, x error plus expr=\thisrow{rIhi}-\thisrow{rI}, x error minus expr=\thisrow{rI}-\thisrow{rIlo}] {data/main.dat};
\addplot+[pt=lockc, mark=*, ci=lockc]
  table[x=rP0, y expr=\thisrow{idx}, x error plus expr=\thisrow{rP0hi}-\thisrow{rP0}, x error minus expr=\thisrow{rP0}-\thisrow{rP0lo}] {data/main.dat};
\addplot[draw=none, forget plot, point meta=explicit,
  nodes near coords={\pgfmathprintnumber[fixed,precision=0]{\pgfplotspointmeta}},
  nodes near coords style={font=\tiny, text=lockc!80!black, anchor=west, xshift=0.5pt, yshift=-0.5pt}]
  table[x=rP0hi, y=idx, meta=rP0] {data/main.dat};
\addplot+[pt=notec, mark=diamond*, mark size=2.8pt, ci=notec]
  table[x=rP1, y expr=\thisrow{idx}+0.22, x error plus expr=\thisrow{rP1hi}-\thisrow{rP1}, x error minus expr=\thisrow{rP1}-\thisrow{rP1lo}] {data/main.dat};
\nextgroupplot[title={Task success (\% of episodes)}, xmin=30, xmax=100, xtick={40,60,80,100}]
\rowbandsfrom{30}{100}
\addplot+[only marks, mark=|, mark size=4.5pt, mark options={line width=0.9pt, draw=infoc}, color=infoc]
  table[x=optI, y=idx] {data/main.dat};
\addplot+[only marks, mark=|, mark size=4.5pt, mark options={line width=0.9pt, draw=ink}, color=ink]
  table[x=optL, y=idx] {data/main.dat};
\addplot+[pt=infoc, mark=square*, mark size=2.1pt, ci=infoc]
  table[x=sI, y expr=\thisrow{idx}-0.22, x error plus expr=\thisrow{sIhi}-\thisrow{sI}, x error minus expr=\thisrow{sI}-\thisrow{sIlo}] {data/main.dat};
\addplot+[pt=lockc, mark=*, ci=lockc]
  table[x=sP0, y expr=\thisrow{idx}, x error plus expr=\thisrow{sP0hi}-\thisrow{sP0}, x error minus expr=\thisrow{sP0}-\thisrow{sP0lo}] {data/main.dat};
\addplot+[pt=notec, mark=diamond*, mark size=2.8pt, ci=notec]
  table[x=sP1, y expr=\thisrow{idx}+0.22, x error plus expr=\thisrow{sP1hi}-\thisrow{sP1}, x error minus expr=\thisrow{sP1}-\thisrow{sP1lo}] {data/main.dat};
\end{groupplot}
\node[anchor=south, font=\scriptsize, text=ink] at ($(group c1r1.north)!0.5!(group c2r1.north)+(0,0.55)$) {%
  \tikz[baseline=-0.6ex]\node[rectangle, fill=infoc, minimum size=5pt, inner sep=0]{}; info\quad
  \tikz[baseline=-0.6ex]\node[circle, fill=lockc, minimum size=5.5pt, inner sep=0]{}; lock\quad
  \tikz[baseline=-0.6ex]\node[diamond, fill=notec, minimum size=6.5pt, inner sep=0]{}; lock+notes\quad
  \tikz[baseline=-0.6ex]\draw[infoc, line width=0.9pt] (0,-0.09) -- (0,0.09);\,\tikz[baseline=-0.6ex]\draw[ink, line width=0.9pt] (0,-0.09) -- (0,0.09); optimal policy (info, lock) on the same hidden worlds};
\end{tikzpicture}
  \caption{\DL worlds (30 worlds, 360 episodes per model and condition). Left: share of episodes in which
  the model reads the trap record; the numbers give the \lock value. Right: task success; the ticks mark the optimal policy's success on
  the same hidden worlds under \info (grey) and \lock (black). Bars are 95\% bootstrap intervals over
  worlds; a cell that is 0\% in every world has none.}
  \label{fig:main}
\end{figure*}

\begin{table*}[t]
  \centering\small
  \setlength{\tabcolsep}{4.5pt}
  \begin{tabular}{@{}l ccc ccc c cc@{}}
\toprule
 & \multicolumn{3}{c}{\DL: trap read} & \multicolumn{3}{c}{\DL: success} & \DL & \multicolumn{2}{c}{\ML (lock)} \\
\cmidrule(lr){2-4}\cmidrule(lr){5-7}\cmidrule(lr){8-8}\cmidrule(l){9-10}
Model & info & lock & +notes & info & lock & +notes & blocked change & key read & side-trap read \\
\midrule
DeepSeek Flash & 36 & 4 & 0 & 94 & 89 & 91 & 22 & 100 & 1 \\
Claude Haiku 5.5 & 43 & 28 & 4 & 94 & 82 & 89 & 5 & 96 & 9 \\
GPT-6 Luna & 48 & 39 & 12 & 94 & 76 & 88 & 2 & 99 & 23 \\
Claude Sonnet 5.5 & 41 & 56 & 0 & 94 & 67 & 91 & 39 & 89 & 24 \\
GPT-4o-mini & 79 & 72 & 81 & 73 & 45 & 39 & 75 & 83 & 27 \\
\midrule
\textit{Optimal policy} & -- & 0 & 0 & 94 & 91 & 91 & -- & 100 & 0 \\
\bottomrule
\end{tabular}

  \caption{Main results (\%). Trap read: episodes with a read of the trap record. Blocked change: among the
  \lock episodes in which the trap read showed the outcome that locks a change the agent needs, the share in
  which the model then attempted that change (pooled over worlds; DeepSeek has \kDeepSeekLockDLHarmEps{}
  such episodes); every such attempt came when no plan could still succeed. Key read: episodes with a read of the key record; side-trap read: episodes with a harmful read of
  another record, over all 15 \ML worlds (12 of which have one). Intervals, regret and the \LA worlds are
  in Appendix~\ref{app:full}.}
  \label{tab:main}
\end{table*}

\subsection{Trap reads}
Figure~\ref{fig:main} and Table~\ref{tab:main} give the main result. Under \lock, DeepSeek Flash reads
the trap in \kDeepSeekLockDLRead{} of 360 episodes and stays close to optimal success. Haiku, Luna,
Sonnet and GPT-4o-mini read it in a quarter to three quarters of episodes. For Haiku, Luna and Sonnet,
success falls below the optimum by about the amount the trap reads cost; GPT-4o-mini loses more, because
it also fails to produce valid actions and makes errors when it travels. A lenient reading of its malformed
travel actions changes its success by at most \nLenFourominiMaxSucc{} points
(Appendix~\ref{app:settings}). Regret under \lock is higher than under \info for Haiku, Luna, Sonnet and GPT-4o-mini in
\cHaikuLockInfoDLRegretUp{}, \cLunaLockInfoDLRegretUp{}, \cSonnetLockInfoDLRegretUp{} and
\cFourominiLockInfoDLRegretUp{} of the 30 worlds respectively, and lower in
\cHaikuLockInfoDLRegretDown{}, \cLunaLockInfoDLRegretDown{}, \cSonnetLockInfoDLRegretDown{} and
\cFourominiLockInfoDLRegretDown. For Haiku, Luna and Sonnet, at least 97\% of the regret under \lock
comes from the reads themselves.

Haiku, Luna and Sonnet play the \info condition with regret below 0.001, so the failure is specific to the
lock. GPT-4o-mini is below the optimum under \info as well (a share of its episodes end without a valid
action) and falls further under \lock.

The rule changes how often the models read the trap, but only partly, and for Sonnet in the wrong
direction. Compared with \info, Haiku, Luna and GPT-4o-mini read the trap less often under \lock
(\cHaikuLockInfoDLRead, \cLunaLockInfoDLRead{} and \cFourominiLockInfoDLRead{} points; Luna's interval
includes zero), DeepSeek much less (\cDeepSeekLockInfoDLRead), and Sonnet more often
(\cSonnetLockInfoDLRead{} points; \cSonnetLockInfoDLReadUp{} worlds up, \cSonnetLockInfoDLReadDown{}
down). Sonnet's replies do not say why: like Haiku's, they mostly present the trap as a way to find out
which change is needed (\S\ref{sec:qualitative}).

For Haiku, Luna and Sonnet, the trap reads are not a misunderstanding of the rule. When a trap read
shows the outcome that locks a change they need, Haiku and Luna almost never attempt that change
afterwards (Table~\ref{tab:main}, \emph{blocked change}); they give up or change something else. Sonnet
attempts it in about two fifths of such episodes, always when no plan could still succeed, and almost
all of these replies say that the change will fail, most adding that trying costs nothing. GPT-4o-mini
attempts it in three quarters of such episodes, and its replies rarely say that it will fail.

Sonnet, a larger model of the same family as Haiku, reads the trap about twice as often. The difference
is not the effort setting: Haiku run at effort high, Sonnet's setting, reads the trap in
\nHaikuHighLockDLRead\% of \lock episodes against \nHaikuLockDLRead\% at effort medium, with the same
success. It may lie in how much the models write before acting. Sonnet produces less than half as many
output tokens per step as Haiku at either setting, and for every model, in most of the worlds where its
first step sometimes reads the trap and sometimes does not, the first steps that read it are shorter. We did not control the length of reasoning and compare
few models, so this is an observation rather than a scaling result, though it resembles tasks on which
larger models do worse \citep{mckenzie2023inverse}.

\subsection{Selectivity}
\label{sec:selectivity}

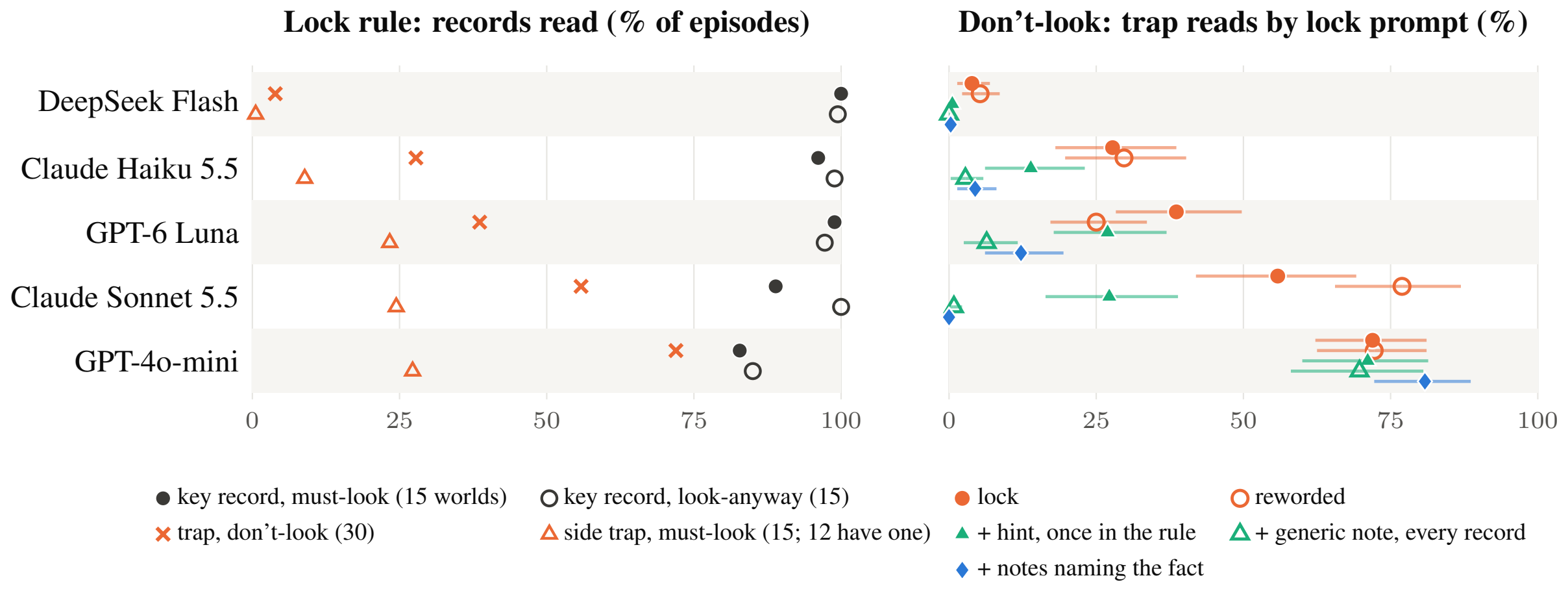
\begin{figure*}[t]
  \centering
\begin{tikzpicture}
\begin{groupplot}[group style={group size=2 by 1, horizontal sep=1.1cm, yticklabels at=edge left},
                  dotplot, width=0.375\textwidth, height=3.4cm]
\nextgroupplot[title={Lock rule: records read (\% of episodes)},
               yticklabels from table={data/main.dat}{label},
               y tick label style={font=\footnotesize, text=ink},
               legend style={at={(0.5,-0.22)}, anchor=north, legend columns=2, draw=none, fill=none,
                             font=\scriptsize, text=ink, /tikz/every even column/.append style={column sep=8pt}},
               legend cell align=left]
\rowbands{100}
\addplot+[only marks, mark=*, mark size=2.2pt, mark options={fill=keyc, draw=white, line width=0.4pt}, color=keyc]
  table[x=ml, y expr=\thisrow{idx}-0.16] {data/selectivity.dat};
\addlegendentry{key record, must-look (15 worlds)}
\addplot+[only marks, mark=o, mark size=2.2pt, mark options={draw=keyc, line width=0.8pt}, color=keyc]
  table[x=ldl, y expr=\thisrow{idx}+0.16] {data/selectivity.dat};
\addlegendentry{key record, look-anyway (15)}
\addplot+[only marks, mark=x, mark size=2.6pt, mark options={draw=lockc, line width=1.1pt}, color=lockc]
  table[x=dl, y expr=\thisrow{idx}-0.16] {data/selectivity.dat};
\addlegendentry{trap, don't-look (30)}
\addplot+[only marks, mark=triangle, mark size=2.4pt, mark options={draw=lockc, line width=0.8pt}, color=lockc]
  table[x=mlo, y expr=\thisrow{idx}+0.16] {data/selectivity.dat};
\addlegendentry{side trap, must-look (15; 12 have one)}
\nextgroupplot[title={Don't-look: trap reads by lock prompt (\%)},
               legend style={at={(0.5,-0.22)}, anchor=north, legend columns=2, draw=none, fill=none,
                             font=\scriptsize, text=ink, /tikz/every even column/.append style={column sep=6pt}},
               legend cell align=left]
\rowbands{100}
\addplot+[pt=lockc, mark=*, ci=lockc]
  table[x=p0, y expr=\thisrow{idx}-0.32, x error plus expr=\thisrow{p0hi}-\thisrow{p0}, x error minus expr=\thisrow{p0}-\thisrow{p0lo}] {data/variants.dat};
\addlegendentry{lock}
\addplot+[only marks, mark=o, mark size=2.3pt, mark options={draw=lockc, line width=0.9pt, fill=white}, color=lockc, ci=lockc]
  table[x=pr, y expr=\thisrow{idx}-0.16, x error plus expr=\thisrow{prhi}-\thisrow{pr}, x error minus expr=\thisrow{pr}-\thisrow{prlo}] {data/variants.dat};
\addlegendentry{reworded}
\addplot+[pt=hintc, mark=triangle*, mark size=2.8pt, ci=hintc]
  table[x=pg, y expr=\thisrow{idx}, x error plus expr=\thisrow{pghi}-\thisrow{pg}, x error minus expr=\thisrow{pg}-\thisrow{pglo}] {data/variants.dat};
\addlegendentry{+ hint, once in the rule}
\addplot+[only marks, mark=triangle, mark size=2.8pt, mark options={draw=hintc, line width=0.9pt}, color=hintc, ci=hintc]
  table[x=pn, y expr=\thisrow{idx}+0.16, x error plus expr=\thisrow{pnhi}-\thisrow{pn}, x error minus expr=\thisrow{pn}-\thisrow{pnlo}] {data/variants.dat};
\addlegendentry{+ generic note, every record}
\addplot+[pt=notec, mark=diamond*, mark size=2.8pt, ci=notec]
  table[x=p1, y expr=\thisrow{idx}+0.32, x error plus expr=\thisrow{p1hi}-\thisrow{p1}, x error minus expr=\thisrow{p1}-\thisrow{p1lo}] {data/variants.dat};
\addlegendentry{+ notes naming the fact}
\end{groupplot}
\end{tikzpicture}
  \caption{Left: under \lock, share of episodes in which each model reads the key record of \ML and \LA
  worlds (filled and open circles), the trap of \DL worlds (crosses), and makes a harmful read of a
  side trap in \ML worlds (triangles). Right: share of \DL episodes with a trap read under five versions of the \lock prompt
  (360 episodes each; bars are 95\% bootstrap intervals over worlds).}
  \label{fig:controls}
\end{figure*}

An agent could avoid traps by reading less. Figure~\ref{fig:controls} (left) shows that this is not what
happens. Under \lock, DeepSeek, Haiku and Luna read the key record of \ML worlds in almost every episode
and Sonnet in \nSonnetLockMLRead\% (it leaves it unread in every episode of one world, in most episodes of
another and in one episode of a third); every model but GPT-4o-mini reads
the key record of \LA worlds, which locks a fact but is worth reading, in nearly every episode. The
models that read traps in \DL worlds also read the side traps of \ML worlds: Luna and Sonnet in about a
quarter of episodes. Haiku, Luna and Sonnet are not reading less under \lock; they are reading without accounting
for what a record locks.

\subsection{Interventions}
\label{sec:variants}

\paragraph{Per-record notes.}
Stating under each record what it locks removes most trap reads for Haiku and Luna and all of them for
Sonnet, and success returns to within four points of the optimum (Table~\ref{tab:main}). The notes do
not reduce reading of key records: in \ML worlds Haiku, Luna and Sonnet read the key record in every
episode, and side-trap reads nearly disappear. GPT-4o-mini is the exception: with notes it reads the trap
more often, not less, and in \ML worlds it reads the key record less and the side trap more; the notes
appear to draw it to the records they describe.

\paragraph{Generic hint and generic note.}
The notes give three things at once: a reminder that reading has consequences, that reminder attached
to each record, and the mapping from each record to the fact it locks. Two variants separate the mapping
from the rest (Figure~\ref{fig:controls}, right). \hint appends one sentence to the lock rule, which is
printed just below the list of records: \emph{``In other words, reading a record can make some changes
impossible.''} The sentence follows from the rule and names no record. It removes a third of Luna's trap
reads and half of Haiku's and Sonnet's, in \cLunaGLockDLReadDown, \cHaikuGLockDLReadDown{} and
\cSonnetGLockDLReadDown{} of 30 worlds. \gnote attaches the same reminder to every record instead
(\emph{``Note: reading this record can make some changes impossible.''}). It names no fact and stands
under records that lock nothing as well, so it does not say which record to avoid. It removes most trap
reads, leaving Haiku at \nHaikuLockNDLRead\%, Luna at \nLunaLockNDLRead\% and Sonnet at
\nSonnetLockNDLRead\%, as much as the notes that name the locked fact (the paired differences, notes
minus generic note, are \cHaikuOneNDLRead,
\cLunaOneNDLRead{} and \cSonnetOneNDLRead{} points, with intervals that include zero). It does not make
these models read less in general: in the \nDLNeededWorlds{} \DL worlds that also contain a record the
agent needs, they read that record in \nHaikuLockNDLNeeded\%, \nLunaLockNDLNeeded\% and
\nSonnetLockNDLNeeded\% of \gnote episodes, about as often as under \lock. In the \DL worlds, the mapping
from record to fact therefore adds nothing that these models cannot work out themselves; what makes the
difference is a reminder attached to each record. The hint, stated once just below the records, helps
only partly.
Whether the repetition or the attachment to each record does the work is not separated. Neither variant
helps GPT-4o-mini. The generic note was not run on the \LA worlds, where the record worth reading itself
locks a fact, so whether it makes models skip such records is not measured.

\subsection{Reworded rule}
\label{sec:r1}
\reworded replaces the lock rule with a paraphrase that keeps its content, avoids its key words and adds
no hint: \emph{``The past always matches what you have seen. Whatever a record
you have read says will still be accurate after you travel back. If one of your changes would make such a
record inaccurate, that change does not take effect, and that part of the past stays as it was.''} Trap
reads remain for every model, and regret under \reworded is higher than under \info for every model
(for DeepSeek, Haiku, Luna, Sonnet and GPT-4o-mini in \cDeepSeekRInfoDLRegretUp, \cHaikuRInfoDLRegretUp,
\cLunaRInfoDLRegretUp, \cSonnetRInfoDLRegretUp{} and \cFourominiRInfoDLRegretUp{} of 30 worlds, with
\cFourominiRInfoDLRegretDown{} worlds the other way for GPT-4o-mini and none for the others). Relative to the original rule, the trap-read rate is unchanged for Haiku, GPT-4o-mini and
DeepSeek, lower for Luna and higher for Sonnet (\cSonnetRLockDLRead{} points). The qualitative result
holds under the rewording; the rates change, so rates from one wording should not be compared across
papers. The paraphrase has one side effect: in \kHaikuRTravelUnread{} \DL travel steps Haiku applies it to
records it has not read and gives up a needed change; with two other travel errors, this raises its travel
regret from 0.003 to 0.027 per episode. A
likely cause is that ``accurate'' echoes the general instruction \emph{``Every record is accurate.''}
Under the paraphrase Haiku also attempts the locked change more often after a trap read that locks a
change it needs (\kHaikuLockRDLBlocked{} of \kHaikuLockRDLHarmEps{} such episodes, against
\kHaikuLockDLBlocked{} of \kHaikuLockDLHarmEps{} under the original rule), again only when no plan could
still succeed. No intervention was tested under the paraphrase.

\subsection{Replies}
\label{sec:qualitative}
Luna's and DeepSeek's replies are almost always the action line alone and give no reasons. In the samples of Haiku's and
Sonnet's replies that we read, their trap reads show the same two patterns. Most
present the record as a way to find out which change is needed, as they would under \info:
\begin{quote}\small
\emph{``Reading the portal status first should show whether the vaccination upload already happened. That
decides which change, if any, is worth making.''} (Sonnet)
\end{quote}
Some also apply the rule correctly to the outcome of the read, as if the read had already happened. In
the world about a grandfather's calligraphy scrolls:
\begin{quote}\small
\emph{``Without reading anything, the best single change is to make the framing happen (success about
0.6, since the wall is booked 6 times in 10). [\ldots] If it shows they weren't, then framing is false
and any change to it would contradict the record, so the original stands and the task fails whatever I
do.''} (Haiku, before reading the framer's photo)
\end{quote}
Haiku has what it needs to decide: travelling now gives 0.6, and reading first gives 0.3, because the
photo shows the scrolls framed only 3 times in 10. It does not compare the two. One of DeepSeek Flash's few replies with reasons, in the
same world, does:
\begin{quote}\small
\emph{``The framer's photo would irrevocably fix whether the scrolls were framed, so reading it can only
reduce our success chance. [\ldots] The best action is to travel back immediately and ensure the
framing, leaving the wall booking to its 60\% chance.''} (DeepSeek Flash)
\end{quote}

\section{Discussion}
One of the five models avoids trap reads without help, and every model but GPT-4o-mini reads the records
it needs. Haiku, Luna and Sonnet understand the rule, as their behaviour after a trap read shows, but do
not apply it before reading. In their replies,
Haiku and Sonnet present the trap as a source of information rather than as an action with a cost;
Luna's replies give no reasons. For these
three models, a sentence that restates the rule's implication removes a third to a half of trap reads
when it stands once in the rule, and most of them when it is attached to every record; in the \DL
worlds, naming the fact each record locks adds nothing beyond that. These models can infer what a record
locks; they do so reliably only when the reminder is attached to each record. No intervention
helps GPT-4o-mini, and the notes make it worse. For deployed agents, this suggests a hypothesis that we
did not test: a reminder of an observation's consequences may work better attached to each tool, for
example in its description, than stated once in the system prompt, although effect hints in function
documentation had no measurable effect in the tool-calling benchmark of \citet{kovalenko2026careful}.

\section*{Limitations}
The worlds are small and stylised, written for the study rather than drawn from real tasks; the time-travel framing
makes the cost of observation explicit, and real settings rarely do. There are 60 worlds and
every interval is over worlds, so effects of a few points are not resolvable. The worlds were drafted
with an LLM (Claude Opus 5.5) and audited by LLM readers of the same model, and two of the evaluated
models are Claude models; human validation of the worlds is planned but not done. The audit's stopping rule was revised three times after seeing results (Appendix~\ref{app:audit}), the fixes made after
the last round were not blind-read, and the residual error rate of the final text is unmeasured. We test one paraphrase of the rule, one generic hint and one wording of the general
instructions; the reworded rule changed the rates, so the rates are specific to the wording. The general
instructions are the same under both rules, but they may interact with the rule's wording, as the
reworded rule shows (\S\ref{sec:r1}), and they may affect the rates. The generic note was tested on the
\DL worlds only: there the models still read the records they need, which lock nothing, but whether it makes models skip a record that locks a fact and is still
worth reading is not measured, and it does not separate repetition from attachment to each record. How
much the models reason before acting was not controlled. The reply patterns in \S\ref{sec:qualitative} come from reading samples of
replies, not from a coded count, and the models' hidden reasoning was not recorded.
GPT-4o-mini fails often for reasons unrelated to the lock rule. A lenient re-score, whose rule was written
after seeing these failures, changes no conclusion (Appendix~\ref{app:settings}), but it recovers only about
half of them.

\section*{Ethics and use of AI}
The study uses no human subjects or personal data. The worlds were drafted with an LLM (Claude Opus 5.5)
in an interactive session under the author's direction and authoring rules, and revised by the author;
the audit's blind readers were runs of the same model; the experiment code and analysis were written
with the same assistant and reviewed by the author; the text of this paper was drafted with the
assistant from the author's outline and results, and edited by the author. All design decisions,
verification and claims are the author's responsibility.

\bibliography{refs}

\clearpage
\appendix

\section{Prompts}
\label{app:prompt}
Figure~\ref{fig:prompt} shows the full \lock prompt of the world in Figure~\ref{fig:teaser} at the first
turn. Under \info, only the paragraph starting \emph{Rule:} differs. Later turns add a list \emph{Records
you have read so far} with the text of each record read. An invalid action is answered by appending
\emph{``Note: $\langle$problem$\rangle$ Please answer again.''} to the prompt, where the problem is, for
example, \emph{``head=left\_behind is not an available change''}. Under \notes, every record carries a note, here \emph{``Note: reading this record locks:
whether the dragon head went on the van.''} under the photo and \emph{``Note: reading this record locks
nothing you could change.''} under the notice and the route. \hint appends to the lock rule: \emph{``In other
words, reading a record can make some changes impossible.''} \gnote keeps the lock rule and puts
\emph{``Note: reading this record can make some changes impossible.''} under every record. \reworded replaces the lock rule with the paraphrase quoted in \S\ref{sec:r1}, preceded by
\emph{``Rule:''}. An earlier draft read as a warning about reading and was replaced before the
runs.

\begin{figure*}[t]
\begin{tcolorbox}[rulebox, fontupper=\scriptsize]
\input{tables/prompt_w15_lock_p0.tex}
\end{tcolorbox}
\caption{The full prompt of the world in Figure~\ref{fig:teaser} under \lock, first turn. The record
descriptions are the frozen world text.}
\label{fig:prompt}
\end{figure*}

\section{Full results}
\label{app:full}
Tables~\ref{tab:full-dl}--\ref{tab:full-la} give every model and condition (the effort control is
reported in the text of \S\ref{sec:results}), with 95\% bootstrap
intervals over worlds for target reads, harmful target reads and success (none for a rate that is 0\% or 100\% in every
world). Target read: any read of the target record. Harmful: a read whose regret exceeds
0.005, of the target or of any other record (for \ML worlds under \lock, \emph{Harmful: other} is the
side-trap read of Table~\ref{tab:main}); -- under \info, where no read is harmful. In \ML and \LA
worlds the target is never harmful by construction. Every cell has 12 episodes per world.

\begin{table}[t]\centering\small
\begin{tabular}{@{}clccc@{}}
\toprule
Round & Threshold in force & Agree & A & B (worlds) \\
\midrule
1 & none yet & 60/60 & -- & -- \\
2 & B $\le$ 3 & 60/60 & 0 & 23 \\
3 & B $\le$ 3 & 60/60 & 1 & 16 \\
4 & B $\le$ 3 & 60/60 & 0 & 14 \\
5 & B $\le$ 10 & 60/60 & 0 & 9 \\
6 & B $\le$ 5 & 60/60 & 0 & 10 \\
7 & B $<$ 10 & 60/60 & 0 & 8 \\
\bottomrule
\end{tabular}
\caption{Blind-reader rounds. In each round, three fresh LLM readers (Claude Opus 5.5) each solved 20 of
the 60 shuffled, anonymised worlds. Agree: solutions that match the solver. Round 1 preceded the A/B
classification; its objections led to fixes in 37 worlds.}
\label{tab:audit}
\end{table}

\begin{table*}[t]\centering\small
\resizebox{\textwidth}{!}{\begin{tabular}{@{}llrrrrrr@{}}
\toprule
Model & Condition & Target read & Harmful: target & Harmful: other & Success & Regret & Model failure \\
\midrule
DeepSeek Flash & \info & 36 [22, 51] & -- & -- & 94 [89, 98] & 0.002 & 0 \\
 & \lock & 4 [1, 7] & 4 [1, 7] & 0 & 89 [83, 95] & 0.016 & 0 \\
 & \hint & 1 [0, 1] & 1 [0, 1] & 0 & 91 [84, 97] & 0.003 & 0 \\
 & \gnote & 0 & 0 & 0 & 91 [84, 97] & 0.000 & 0 \\
 & \reworded & 5 [2, 9] & 5 [2, 8] & 0 & 90 [83, 96] & 0.020 & 0 \\
 & \notes & 0 [0, 1] & 0 [0, 1] & 0 & 91 [84, 96] & 0.002 & 0 \\
\addlinespace[2pt]
Claude Haiku 5.5 & \info & 43 [29, 57] & -- & -- & 94 [89, 98] & 0.000 & 0 \\
 & \lock & 28 [18, 39] & 27 [17, 38] & 0 & 82 [74, 89] & 0.107 & 0 \\
 & \hint & 14 [6, 23] & 14 [6, 23] & 0 & 86 [77, 94] & 0.049 & 0 \\
 & \gnote & 3 [0, 6] & 3 [0, 6] & 0 & 89 [82, 95] & 0.021 & 0 \\
 & \reworded & 30 [20, 40] & 29 [20, 40] & 0 & 76 [68, 84] & 0.160 & 1 \\
 & \notes & 4 [1, 8] & 4 [1, 7] & 0 & 89 [83, 95] & 0.019 & 0 \\
\addlinespace[2pt]
GPT-6 Luna & \info & 48 [36, 60] & -- & -- & 94 [89, 98] & 0.000 & 0 \\
 & \lock & 39 [28, 50] & 38 [28, 49] & 0 & 76 [67, 84] & 0.170 & 0 \\
 & \hint & 27 [18, 37] & 26 [17, 36] & 0 & 80 [71, 88] & 0.112 & 0 \\
 & \gnote & 6 [3, 12] & 6 [2, 11] & 0 & 89 [81, 96] & 0.022 & 0 \\
 & \reworded & 25 [17, 34] & 24 [16, 33] & 0 & 79 [70, 86] & 0.118 & 0 \\
 & \notes & 12 [6, 19] & 12 [6, 19] & 0 & 88 [80, 94] & 0.049 & 0 \\
\addlinespace[2pt]
Claude Sonnet 5.5 & \info & 41 [26, 57] & -- & -- & 94 [89, 98] & 0.000 & 0 \\
 & \lock & 56 [42, 69] & 55 [42, 68] & 0 & 67 [57, 77] & 0.242 & 0 \\
 & \hint & 27 [16, 39] & 27 [16, 39] & 0 & 78 [68, 87] & 0.107 & 0 \\
 & \gnote & 1 [0, 2] & 1 [0, 2] & 0 & 91 [84, 97] & 0.003 & 0 \\
 & \reworded & 77 [66, 87] & 77 [65, 87] & 0 & 51 [42, 61] & 0.366 & 0 \\
 & \notes & 0 & 0 & 0 & 91 [84, 97] & 0.000 & 0 \\
\addlinespace[2pt]
GPT-4o-mini & \info & 79 [69, 89] & -- & -- & 73 [64, 82] & 0.216 & 13 \\
 & \lock & 72 [62, 81] & 69 [58, 79] & 1 & 45 [36, 53] & 0.449 & 12 \\
 & \hint & 71 [60, 81] & 70 [59, 80] & 1 & 47 [38, 55] & 0.424 & 11 \\
 & \gnote & 70 [58, 81] & 67 [55, 78] & 1 & 47 [38, 56] & 0.416 & 10 \\
 & \reworded & 72 [63, 81] & 70 [59, 79] & 1 & 50 [42, 58] & 0.399 & 6 \\
 & \notes & 81 [72, 89] & 79 [70, 87] & 1 & 39 [30, 48] & 0.503 & 13 \\
\bottomrule
\end{tabular}
}
\caption{Full results (\%; regret in probability units), \DL worlds (30 worlds).}
\label{tab:full-dl}
\end{table*}
\begin{table*}[t]\centering\small
\resizebox{0.9\textwidth}{!}{\begin{tabular}{@{}llrrrrr@{}}
\toprule
Model & Condition & Target read & Harmful: other & Success & Regret & Model failure \\
\midrule
DeepSeek Flash & \info & 91 [74, 100] & -- & 98 [95, 100] & 0.000 & 0 \\
 & \lock & 100 & 1 & 98 [93, 100] & 0.004 & 0 \\
 & \hint & 100 & 0 & 98 [93, 100] & 0.000 & 0 \\
 & \reworded & 100 & 2 & 97 [92, 100] & 0.012 & 0 \\
 & \notes & 100 & 0 & 98 [93, 100] & 0.000 & 0 \\
\addlinespace[2pt]
Claude Haiku 5.5 & \info & 91 [77, 100] & -- & 98 [95, 100] & 0.000 & 0 \\
 & \lock & 96 [91, 100] & 9 & 94 [88, 99] & 0.048 & 1 \\
 & \hint & 97 [92, 100] & 6 & 94 [86, 100] & 0.018 & 0 \\
 & \reworded & 94 [87, 100] & 13 & 90 [81, 97] & 0.071 & 0 \\
 & \notes & 100 & 0 & 98 [93, 100] & 0.000 & 0 \\
\addlinespace[2pt]
GPT-6 Luna & \info & 88 [70, 100] & -- & 98 [95, 100] & 0.000 & 0 \\
 & \lock & 99 [97, 100] & 23 & 86 [76, 94] & 0.136 & 0 \\
 & \hint & 99 [97, 100] & 12 & 89 [82, 96] & 0.071 & 1 \\
 & \reworded & 99 [97, 100] & 16 & 90 [84, 96] & 0.103 & 1 \\
 & \notes & 100 & 3 & 96 [91, 99] & 0.020 & 0 \\
\addlinespace[2pt]
Claude Sonnet 5.5 & \info & 91 [76, 100] & -- & 98 [95, 100] & 0.000 & 0 \\
 & \lock & 89 [72, 100] & 24 & 86 [77, 93] & 0.110 & 0 \\
 & \hint & 89 [73, 100] & 14 & 93 [84, 99] & 0.045 & 0 \\
 & \reworded & 90 [73, 100] & 38 & 78 [65, 90] & 0.207 & 0 \\
 & \notes & 100 & 0 & 98 [93, 100] & 0.000 & 0 \\
\addlinespace[2pt]
GPT-4o-mini & \info & 91 [83, 98] & -- & 86 [77, 93] & 0.109 & 4 \\
 & \lock & 83 [67, 96] & 27 & 75 [62, 86] & 0.230 & 2 \\
 & \hint & 84 [72, 95] & 26 & 78 [69, 87] & 0.219 & 3 \\
 & \reworded & 86 [73, 97] & 28 & 74 [63, 84] & 0.229 & 1 \\
 & \notes & 67 [53, 81] & 46 & 64 [53, 75] & 0.363 & 5 \\
\bottomrule
\end{tabular}
}
\caption{Full results (\%; regret in probability units), \ML worlds (15 worlds, 12 with a side trap).}
\label{tab:full-ml}
\end{table*}
\begin{table*}[t]\centering\small
\resizebox{0.9\textwidth}{!}{\begin{tabular}{@{}llrrrrr@{}}
\toprule
Model & Condition & Target read & Harmful: other & Success & Regret & Model failure \\
\midrule
DeepSeek Flash & \info & 100 & -- & 92 [83, 98] & 0.000 & 0 \\
 & \lock & 99 [98, 100] & 1 & 89 [82, 96] & 0.010 & 0 \\
 & \hint & 100 & 1 & 91 [83, 97] & 0.004 & 0 \\
 & \reworded & 99 [97, 100] & 1 & 89 [82, 95] & 0.014 & 0 \\
 & \notes & 99 [98, 100] & 0 & 91 [83, 98] & 0.001 & 0 \\
\addlinespace[2pt]
Claude Haiku 5.5 & \info & 99 [97, 100] & -- & 91 [82, 98] & 0.003 & 0 \\
 & \lock & 99 [97, 100] & 1 & 90 [83, 96] & 0.012 & 0 \\
 & \hint & 100 & 0 & 90 [81, 97] & 0.006 & 1 \\
 & \reworded & 98 [95, 100] & 6 & 83 [74, 92] & 0.061 & 0 \\
 & \notes & 100 & 0 & 91 [83, 98] & 0.000 & 0 \\
\addlinespace[2pt]
GPT-6 Luna & \info & 99 [98, 100] & -- & 93 [86, 98] & 0.001 & 0 \\
 & \lock & 97 [93, 100] & 17 & 77 [64, 89] & 0.122 & 0 \\
 & \hint & 98 [96, 100] & 12 & 81 [71, 91] & 0.090 & 1 \\
 & \reworded & 98 [97, 100] & 13 & 81 [68, 91] & 0.115 & 2 \\
 & \notes & 97 [94, 100] & 3 & 88 [79, 95] & 0.032 & 0 \\
\addlinespace[2pt]
Claude Sonnet 5.5 & \info & 100 & -- & 91 [82, 98] & 0.000 & 0 \\
 & \lock & 100 & 3 & 89 [82, 96] & 0.023 & 0 \\
 & \hint & 100 & 1 & 91 [83, 97] & 0.004 & 0 \\
 & \reworded & 100 & 12 & 83 [74, 92] & 0.083 & 0 \\
 & \notes & 100 & 0 & 91 [83, 98] & 0.000 & 0 \\
\addlinespace[2pt]
GPT-4o-mini & \info & 94 [88, 98] & -- & 78 [67, 89] & 0.147 & 8 \\
 & \lock & 85 [75, 93] & 10 & 72 [59, 84] & 0.200 & 7 \\
 & \hint & 88 [79, 96] & 13 & 66 [52, 79] & 0.232 & 8 \\
 & \reworded & 90 [83, 96] & 9 & 68 [56, 80] & 0.213 & 8 \\
 & \notes & 93 [90, 97] & 11 & 72 [62, 81] & 0.177 & 6 \\
\bottomrule
\end{tabular}
}
\caption{Full results (\%; regret in probability units), \LA worlds (15 worlds, 3 with a side trap).}
\label{tab:full-la}
\end{table*}

\section{Audit rounds}
\label{app:audit}
Table~\ref{tab:audit} lists every blind-reader round; A and B are as defined in \S\ref{sec:audit}, and B
counts the worlds affected. The threshold in force in each round is given in the table; it was set after
round 1 and revised after rounds 4, 5 and 6, each time before the next round and after seeing the results
so far. From round 7 on the rule was to continue until met. The first rule also allowed at most four rounds,
and round 5 was announced as the last; both limits were dropped when the rule was revised. The four problems introduced by the author's
own fixes were found in rounds 3 (one, the only A), 5 (one) and 6 (two). The fixes made after round 7
were not blind-read.

\section{Run settings and costs}
\label{app:settings}
Claude models: \texttt{claude-haiku-5-5} (effort medium; effort high for the effort control) and
\texttt{claude-sonnet-5-5} (effort high),
streamed, at most 32,000 output tokens per call, adaptive thinking (the default; no thinking parameter
was sent). OpenAI: \texttt{gpt-4o-mini} (served as \texttt{gpt-4o-mini-2024-07-18}) and
\texttt{gpt-6-luna} (no dated snapshot is reported by the API). DeepSeek: \texttt{deepseek-flash}, which
runs in thinking mode by default (no dated snapshot). The harness recorded the visible text of each
reply, not the models' hidden reasoning. All runs took place on 25--26 November 2026 (UTC).
Episodes: 12 per world and condition; \info, \lock, \notes, \hint and \reworded on all 60 worlds, \gnote
and the effort control (\info and \lock) on the 30 \DL worlds. API spend: Haiku \$5.21 (including
the effort control), Sonnet \$45.02, GPT-4o-mini \$2.01, GPT-6 Luna \$1.78, DeepSeek Flash \$13.94
(off-peak pricing except the generic-note run); \$67.96 in total. No episode was lost to an API error (failed calls were retried). Model failures: in most of
GPT-4o-mini's, the last of the three replies writes a change with a value that is not on the list (for
example \texttt{basketed=true} for \texttt{basketed=basketed}) or puts a change among the choices; GPT-6 Luna failed 8 of 1,440 \hint and
\reworded episodes by copying the trailing ``, \ldots'' of the action template three times, and
excluding them changes its success by about one point at most; Haiku had two refusals (one world
under \reworded) and three episodes without a valid action, and one more refusal in the effort control.
As a check, every episode of every model was re-scored with a lenient reading of travel actions. When the first reply in a turn was a travel action that
was invalid only in form, the plan the model evidently meant was scored: a value written in another form
(\texttt{basketed=true} when \texttt{basketed} is the only available value, or the value of a change or a
choice followed by extra text), a change written among the choices or the reverse, or the copied
``, \ldots''. Replies with more than one action are not re-read. A travel ends the
episode, so this needs no simulation. The rule was written after seeing GPT-4o-mini's failures. Under it,
GPT-4o-mini's success changes by at most \nLenFourominiMaxSucc{} points in any family and condition and its
trap reads by at most \nLenFourominiMaxRead{} points, and its gap between \info and \lock in the \DL worlds is
unchanged; \kLenFourominiValid{} of its \kLenFourominiFail{} model failures become valid actions,
\kLenFourominiWin{} of which succeed. The rule does not cover the rest, such as a fact name and its value
swapped, an invented fact name, a missing part of the action line or a re-read, and it does not re-read the
later replies of a turn, which answer the harness's feedback. Luna's success changes by at most \nLenLunaMaxSucc{} points, Haiku's by less than one,
and Sonnet's and DeepSeek's not at all. No conclusion changes.
Pilot runs on earlier versions of the text (Haiku, GPT-4o-mini, DeepSeek) found wording problems
and are not reported.

\end{document}